\documentclass[manuscript,sigconf]{acmart} 
\AtBeginDocument{%
  }

\setcopyright{cc}
\copyrightyear{2025}
\acmYear{2025}
\acmConference[STAIG'25]{CHI2025 Workshop, Sociotechnical AI Governance}{April 27, 2025}{Yokohama, Japan}

\newif\ifdraft
 \drafttrue
 \draftfalse
\ifdraft
  \newcommand{\todo}[1]{\textsf{\textbf{\textcolor{red}{[TODO: #1]}}}}
  \newcommand{\key}[1]{\textsf{{\textit{\textcolor{red}{#1:}}}}}
  \newcommand{\yuri}[1]{\textsf{\textcolor{orange}{\textbf{Yuri:} \textit{#1}}}}
\else
  \newcommand{\todo}[1]{}
  \newcommand{\key}[1]{}
  \newcommand{\yuri}[1]{}
\fi

\begin{document}

\title{Accountability of Generative AI: Exploring a Precautionary Approach for ``Artificially Created Nature''}

\author{Yuri Nakao}
\email{nakao.yuri@fujitsu.com}
\affiliation{%
  \institution{Fujitsu Limited}
  \city{Kawasaki-city}
  \state{Kanagawa}
  \country{Japan}
}
\affiliation{%
  \institution{The University of Tokyo}
  \city{Meguro-ku}
  \state{Tokyo}
  \country{Japan}
}


\begin{abstract}
  The rapid development of generative artificial intelligence (AI) technologies raises concerns about the accountability of sociotechnical systems. Current generative AI systems rely on complex mechanisms that make it difficult for even experts to fully trace the reasons behind the outputs. This paper first examines existing research on AI transparency and accountability and argues that transparency is not a sufficient condition for accountability but can contribute to its improvement. We then discuss that if it is not possible to make generative AI transparent, generative AI technology becomes ``artificially created nature'' in a metaphorical sense, and suggest using the precautionary principle approach to consider AI risks. Finally, we propose that a platform for citizen participation is needed to address the risks of generative AI.
\end{abstract}



\keywords{Generative AI, Accountability, Transparency, Precautionary approach}


\maketitle
\section{Introduction}
The technologies of generative artificial intelligence (AI) have brought issues related to the accountability of socio-technical systems. Current generative AI systems, such as those that output text and images, work based on, e.g., diffusion models~\cite{ho2020denoising} and transformers based on attentional mechanisms~\cite{vaswani2017attention}. On the other hand, these technologies are complex, and it is difficult to fully trace why a specific output was generated. For example, the attention mechanism dynamically changes the weights between a specific word and other words in a document when calculating what words to present after the specific word. Because of this dynamic behavior, there is still no definitive technology to fully visualize the behavior of the attention mechanism. In addition, prompt engineering research is increasingly being conducted by accessing the black-box generative AI systems of some companies through APIs. This kind of research can progress even if experts do not have direct access to the mechanisms of the technology, such as algorithms or network structure~\cite{liao2024ai}. On the other hand, when the results from a technology whose mechanism is not clear change, it is difficult to determine whether the change is the result of prompt engineering or a change in the mechanism by the company that manages the generative AI system. This makes it more unclear where responsibility lies for changes in the results from the system. This use of technology with little or no traceability leads to a situation where even the designers and developers of the system cannot guarantee the results from the technology, reducing the accountability of the decision-making process in which the technology is included.

In this paper, we first examine the assumptions about AI transparency and accountability from existing research and discuss how transparency of AI technology is not a sufficient condition for accountability but does contribute to improving accountability. We will then discuss, based on the precautionary principle, how we should think about the risks posed by generative AI technology if it proves inherently difficult to ensure transparency in generative AI. In this discussion, We metaphorically refer to technologies such as generative AI, in which even experts cannot accurately trace the internal processes, as ``artificially created nature,'' and suggest the need to determine the risks in a way that allows citizens to participate.

\section{Transparency and Accountability of AI}
Existing research on accountability in AI has pointed out that accountability should be ensured throughout the decision-making lifecycle using AI and that transparency of technology is not necessarily helpful for accountability. Novelli et al. defined accountability in AI as an answerability relationship between agents, including natural and legal persons, and argued that goal-based analysis is useful for AI policy-making related to accountability~\cite{novelli2024accountability}. The agents involved in the processes in which AI systems are included are distributed throughout the AI lifecycle. For example, the EU AI-ACT divides the stakeholders into, for example, providers, deployers, importers, and distributors of AI systems~\cite{EUAIACTArticle3}. These stakeholders are distributed throughout the lifecycle of AI systems, including the stages of design, development, and operation. The accountability of AI cannot be addressed only within one particular point in the lifecycle; it needs to be ensured in the interaction among the agents distributed throughout the lifecycle.

Regarding transparency and accountability in AI, there are existing studies on the technical possibilities of transparency~\cite{liao2024ai} and the relationship between transparency and accountability in the field of science, technology, and society~\cite{ananny2018seeing}. From a technical point of view, Liao and Vaughan, for example, discuss the direction of technology to ensure the transparency of large language models (LLMs) in terms of model reporting, publishing evaluation results, providing explanations, and communicating uncertainty~\cite{liao2024ai}. On the other hand, it is also argued that transparency does not lead directly to accountability. Ananny and Crowford list 10 limitations of transparency with regard to algorithmic accountability and argue that attempts to increase transparency may even have a negative impact on ensuring accountability, e.g., when transparency is used to reveal the efforts of marginalized people to rebel against power~\cite{ananny2018seeing}.

While there is some debate about whether transparency contributes to accountability, it is preferable to have transparency in technology when people and organizations take responsibility for the results of the technology. Current generative AI technologies behave in complex ways, such as dynamically changing internal weightings based on input~\cite{vaswani2017attention}, making it difficult for even experts to clarify the reasons for the outputs completely. This means that there is essentially no one who can be accountable for the outputs of the technology in the sense of being able to explain why specific results are provided by the technology. To fully trace and explain the output of a technology, the technology needs to be interpretable and completely transparent. While the transparency of technology is not a sufficient condition for accountability, technologies that increase the transparency of technology, such as technologies for interpretability and explainability, can contribute to increasing accountability.

\section{Transparency of Generative AI and Usage Policy}
While technology transparency is potentially useful in improving accountability, there is currently no definitive technology that makes generative AI interpretable and explainable. If a technology emerges that makes generative AI interpretable or explainable, we should pursue AI accountability as we normally do with systems that can be made transparent, with attention to the points made in existing research. In other words, we should aim for an accountable system in which it is clear in principle who, where, and how to provide the necessary explanations and ensure that the technology behaves as it should.

However, if it becomes clear that it is essentially impossible to make generative AI technology interpretable or accountable, how should we think about the accountability of systems that contain generative AI?

If the technology to make generative AI transparent is essentially unfeasible, there are two possible ways to deal with generative AI:  One is to prohibit its use, and the other is to continue its use. A possible rationale for prohibiting the use of generative AI is that technology should not be used if no one can be held accountable for its results. When the technology itself cannot take responsibility, technology for which responsibility is not obvious should not be included in decisions that are public or high-risk. Although a complete prohibition may be difficult because of the possibility of hidden use of the technology, prohibiting it has certain effects. 

On the other hand, we can choose to continue to use the technology. A possible rationale for continued use is that people can interpret generative AI as a technology that can be under human control and is safe, with some assurance of the relationship between inputs and outputs, even if the internal mechanisms are not entirely clear. This is a view that is common to the use of many technologies that exploit natural phenomena. While humans currently use many technologies that utilize natural phenomena to sustain society, they do not fully understand the mechanisms of all of these natural phenomena. The option of continuing to use generative AI even when it cannot be made transparent is based on the thinking that recognizes generative AI as an ``artificially created nature'' in a metaphorical sense.

\section{Exploration of Precautionary Approach of ``Artificially Created Nature''}

In the following, we discuss the direction of risk coping when using this metaphorical ``artificially created nature'' based on a discussion of the precautionary principle. The precautionary principle states that if there is a risk of harm to the environment or human health, it is necessary to take precautionary measures against such activities as technological innovation, even if the causes have not been adequately proven scientifically~\cite{cameron1991precautionary}. For example, in the case of pollution from a chemical plant emitting toxic substances, even if the risk to human health and the environment is not known at first, if harmful changes are observed in the surrounding environment and residents, the plant operation should be stopped immediately even if there is no scientific basis. On the other hand, in the case of newly developed medicines, which clearly affect the human body, new technologies (i.e., medicines) should not be disseminated until their safety for people is scientifically proven through clinical trials to be assured. These courses of action are a principle or an approach to avoid harming human society and the Earth's nature through human use of nature that we do not fully understand.

There is no societal agreement on how far to apply this precautionary principle to generative AI. Originally, information technology has been rarely incomprehensible to humans. This is because many information technologies have been based on algorithms, computational procedures, and processing protocols at the core of the technology, and it has been possible for humans to trace their behavior. For AI technologies as well, until the 2010s, when the impact of deep learning became known, the common approach was to try to explain the reasons for the processing results of the technology in a way that was understandable to humans~\cite{xu2019explainable}. With regard to generative AI technology, an ``artificially created nature'' in the metaphorical sense, which means the technology created by humans but not entirely understandable by humans, people do not yet have an agreement on what level of safety should be ensured and what level of risk is acceptable, unlike the examples of pharmaceuticals or chemical plants.

On the other hand, the concept of the precautionary principle has the potential to provide a different perspective on accountability for AI technologies that are difficult to be fully transparent. To ensure the accountability of AI systems, the regulatory authorities or governments are currently taking a risk-based governance approach~\cite{Japan2025Interim,G72023Hiroshima,ebers2024truly}. They are trying to make rules for each use case or application corresponding to their risks. While the risk-based approach regulation is rational, the approach has difficulties in evaluating the general risk of the technology itself because the evaluation of the specific application can address only the specific aspect of the technology.  To discuss the essential societal risk of technology and how to regulate it, it is necessary to start to work towards a common agreement on the risk of technology itself.
For example, for nuclear energy, in addition to regulations for each use case, such as weapons~\cite{United2012Treaty}, there are basic agreements on, e.g., how it should be managed and how waste should be handled~\cite{IAEA2006Fundamental}. 
For the generative AI technology, no such agreements exist, and the need for such agreements may not even be recognized. If there is a possibility that the technology poses risks, it is necessary to introduce not only the current idea of accountability based on the risk-based approach but also the more general idea of handling the risks of natural phenomena, i.e., the precautionary principle, into generative AI technology by considering it as ``artificially created nature.''

In order to develop a societal consensus on precautionary principles regarding generative AI technologies, we would like to suggest the need for a platform for public participatory discussion of generative AI. Initial discussions on the safety and risks of AI were mainly conducted by a group of experts~\cite{AIHLEGMember}. Although currently there are multi-stakeholder discussions~\cite{EAIA}, the participation of non-expert, non-representative citizens of any organization is limited. This is possibly because it is difficult for non-expert citizens to gain immediate and accurate knowledge about the technology and because there is little immediate risk of harm to the human body and, therefore, no real feeling of it. However, the difficulty of understanding is the same for other sciences and technologies, and the opportunity to choose what risks are acceptable and what are not should be open to all people. Therefore, it is necessary to design a platform to increase understanding of technology and, at the same time, to explain and discuss the benefits and risks of technology. For this purpose, traditional citizen participation activities such as consensus conferences and citizen juries can be used. In addition, the use of online discussion platforms or the establishment of new ones would be beneficial.

\section{Conclusion}
In this paper, we argue the following: 
\begin{itemize}
    \item The transparency of generative AI can contribute to improving accountability,
    \item If technologies to ensure transparency of generative AI are developed, AI accountability should be ensured in the same way as for conventional AI technologies that can be transparent,
    \item If it turns out that the transparency of generative AI technology is not possible, the risks should be considered based on the precautionary principle,
    \item In the case that generative AI technology is metaphorical ``artificially created nature,'' it is necessary to discuss risk perception through public participation, as with conventional natural science-based technology.
\end{itemize}

Viewing the process by which AI becomes more complex and more difficult for humans to understand as the process by which artifacts become more like natural objects helps discuss the concept of risk for future technologies. This way of thinking will contribute to aligning the risk assessment and use of AI with human values in the future.

\begin{acks}
This work was supported by JST, ACT-X Grant Number JPMJAX21AJ, Japan.
\end{acks}

\bibliographystyle{ACM-Reference-Format}
\bibliography{sample-base}


\end{document}